# Implementation and Assessment of Machine Learning Models for Forecasting Suspected Opioid Overdoses in Emergency Medical Services Data

**Aaron D. Mullen, B.S.[1], Daniel R. Harris, Ph. D.[1], Peter Rock, MPH[1,2], Katherine Thompson Ph. D.[3], Svetla Slavova Ph. D.[4], Jeffery Talbert, Ph. D.[1], V.K. Cody Bumgardner, Ph. D.[1]**

**[1]Institute for Biomedical Informatics, University of Kentucky, Lexington, KY, USA**
**[2]Substance Use Priority Research Area, University of Kentucky, Lexington, KY, USA**
**[3]Dr. Bing Zhang Department of Statistics, University of Kentucky, Lexington, KY, USA**
**[4]Department of Biostatistics, University of Kentucky, Lexington, KY, USA**

**Abstract**

*We present efforts in the fields of machine learning and time series forecasting to accurately predict counts of future suspected opioid overdoses recorded by Emergency Medical Services (EMS) in the state of Kentucky. Forecasts help government agencies properly prepare and distribute resources related to opioid overdoses. Our approach uses county and district level aggregations of suspected opioid overdose encounters and forecasts future counts for different time intervals. Models with different levels of complexity were evaluated to minimize forecasting error. A variety of additional covariates relevant to opioid overdoses and public health were tested to determine their impact on model performance. Our evaluation shows that useful predictions can be generated with limited error for different types of regions, and high performance can be achieved using commonly available covariates and relatively simple forecasting models.*

## 1. Introduction

Opioid use disorder (OUD) remains a persistent public health crisis and epidemic. In 2023, Kentucky had the fifth-largest drug overdose fatality rate in the United States, and 79% of those deaths involved opioid substances[1]. Community-level heterogeneity in OUD and overdose trends complicates state-level responses, suggesting a need for interventions guided by local data[2]. Efforts to reduce opioid overdoses and support opioid overdose control and prevention are limited by delays in data availability and fragmented data systems. In the context of the dynamically changing opioid epidemic, agencies and organizations responsible for monitoring and improving the health of the population need timely data to make critical decisions on resource allocation and targeted responses[3,4].

Time series forecasting can help address the opioid epidemic by predicting both when and where increases or decreases in overdoses are expected to happen. With this knowledge, interventions such as naloxone can be distributed to these areas efficently[5]. Agencies for law enforcement and emergency medical services (EMS) could better anticipate spikes and prepare accordingly, making treatment more accessible and quicker for overdose victims in those areas. Understanding the relationship between opioid overdose events and measures for evidence-based practices for opioid overdose prevention and control will allow development of prediction models that account for current community resources and social determinants of health[6]. State and local stakeholders could plan programs and policies that impact these factors and ideally reduce either the frequency or lethality of opioid overdose events[4,7,8].

The techniques used to forecast opioid overdoses vary depending on factors such as scale, forecasting window, and data availability. Some existing work focuses on individual risk for overdose using statistical or basic machine learning models[9,10], while other studies take a national scale to forecast weekly counts[11]. There is a variety of existing work focused on forecasting overdose counts at different scales[12,13], but this work differs by assessing various geographic and temporal scales across multiple models to determine the optimal data configuration and forecasting models that minimize error and maximize accuracy. Additionally, we discuss the use of novel data sources as covariates to determine their impact on forecasting accuracy.

This article describes approaches to the implementation and evaluation of forecasting models for suspected opioid overdoses (SOOs) requiring EMS response in Kentucky, using different model architectures and covariate data to improve prediction performance. The work is part of a new analytical system, Rapid Actionable Data for Opioid Response in Kentucky (RADOR-KY), which provides state agencies and local stakeholders with actionable timely information to support opioid overdose prevention, harm reduction, evidence-based treatment, and recovery.

## 2. Methods

Time series forecasting is the process of analyzing temporal data to create future predictions, which can be done with simpler, regression-type models that take a statistical approach to analyzing the trends and changes in the data[14], or with more complex machine learning models[15]. These machine learning models are less interpretable and more complicated to build, but they can also outperform statistical approaches, especially with larger sample sizes, increased complexity of data trends, and higher counts of covariates to interpret[16]. These models can identify components of the time series such as the overall trend, contributions from seasonality, or repeating cycles seen throughout the data. Through the identification of these additional components, more can be learned about trends observable from the series.

RADOR-KY forecasts the number of SOOs recorded by EMS agencies in Kentucky using historical data from January 2018 to present day in partnership with the Kentucky Board of Emergency Medical Services. This data source is updated weekly, which enables us to rapidly update and adjust our forecasting efforts. It consists of individual records of every suspected opioid overdose encounter recorded by the EMS agencies, which includes diagnosis information, dates, and address information about where the SOO occurred; we geocoded these addresses to assign each SOO to a county. We aggregated these data into appropriate temporal and geographical levels, which supported the creation of statistical and machine learning models needed to perform the forecasting. To improve our forecasting efforts, we also aggregated covariate data from a variety of data sources related to overdoses or public health.

### 2.1. Geography/Time

The EMS data for SOOs supported geographical aggregation at the state, area development district (ADD), county, zip code, census tract, block group, or block levels. ADDs are specific regions of Kentucky where coordination of programs and services across local county governments may occur to support economic and public health initiatives. Census tracts, block groups, and block levels are regions designated by the United States Census Bureau. The SOO totals are aggregable temporally at a weekly, monthly, quarterly, or yearly interval. We evaluated the sparseness of each geographic region at each interval by measuring the proportion of records having no SOO events.

### 2.2. Models

Different models were evaluated to determine the optimal choice for forecasting. The only limitation on model choice was that all models needed to be able to support past, future, and static covariates.

Three primary models were evaluated for forecasting the number of EMS runs with SOOs for each timestamp and geographical region: Linear Regression[17], the N-Linear LSTF model[18], and the Temporal Fusion Transformer (TFT)[19]. Linear regression is the simplest of these models, using lagged values of the target series and covariate values. This technique is not specific to time series forecasting and was chosen for testing as a comparison against other machine learning models. The N-Linear model is a simple, one-layer neural network and the Temporal Fusion Transformer (TFT) is a deep learning model alternative; both models were designed specifically for time series forecasting. To best capture long-range dependencies, TFT, like other transformer models, consists of multiple layers and uses a self-attention mechanism to weigh the importance of data at different time steps relative to each other.

All three of these models perform well with time series data and possess multivariate capabilities, allowing for multiple distinct time series to be trained on at the same time. This is important for our use case, as each county or geographical region has its own time series which can be evaluated separately. Using all series at once rather than training individual models for each allows the models to capture common trends across the board, while training individual models for each series lacks this collectiveness.

### 2.3. Covariates

Data sources relevant to public health and overdoses were used to create covariates that could aid the model in generating predictions. The EMS SOO data was used to create additional covariates related to overdose location. A 'nearby counties' covariate was created for each county by averaging the EMS SOO totals for other counties in the same ADD, and for ADD predictions, the other adjacent ADDs were averaged instead. Similarly, a 'similar counties' covariate was created by averaging EMS totals for counties with similar trends to the target county. This was done to account for relationships in the trends between counties that may not necessarily be geographically close to one another. To create this 'similar counties' covariate, correlation analysis was performed to determine the linear relationship between the target counties' SOO counts and the totals of other counties lagged by one month; the top five highest absolute value correlations were averaged to create the value for each county. The lagged series are used to identify counties whose trends tend to precede those of the target county. This calculation is only done at the county level and is averaged across counties to obtain ADD level values.

We used the OpenMeteo weather API to gather weather data (temperature, precipitation, sunshine, etc.) for all regions and counties in the state[20]. For each county, social determinants of health (SDOH) data related to unemployment rate, proportion of cost-burdened housing, vehicle access, and municipal housing rates were aggregated per region[21]. For our models, SDOH data act as static covariates that do not change, rather than dynamic covariates available at weekly and monthly intervals. These static covariates are used to improve model performance by distinguishing between different areas with variance in key SDOH characteristics, but they are not updated between timesteps. Medicaid data is provided by the Kentucky Department for Medicaid Services and includes the number of individuals diagnosed with OUD receiving Medications for Opioid Use Disorder (MOUD) and the number of individuals diagnosed with OUD receiving behavioral health treatment. Drug seizure data was provided by the Kentucky State Police and includes where drugs were seized and what substances tested positive in the seizure, which was combined into counts for each geographic region and limited only to opioid substances. In partnership with the Kentucky Department of Corrections (DOC), we used intake and release data, where intakes captured offenders who have a new period of supervision with DOC Probation and Parole and the releases capture offenders who were released from DOC custody. Individuals were assessed and assigned a substance use risk on a scale of 1 to 5; we calculated a weighted average of risk levels for each county and interval for both releases and intakes.

Urine drug testing positivity rates for specimens collected at substance use disorder treatment clinics in Kentucky were provided by Millennium Health. Integrated Nested Laplace Approximations (INLA)-adjusted detection rates were calculated for each county and multiple types of substances, including fentanyl, heroin, prescription opioids, and more. We specifically selected heroin and prescription opioids as covariates after initial testing demonstrated these tests were the best performing. The locations of substance use treatment centers were incorporated as a static covariate. Find Help Now Kentucky[22] provided information about each treatment center and its maximum capacity, which was used to calculate the total number of treatment slots available in each geographic region. We also developed a covariate representing the number of individuals receiving buprenorphine products that are FDA approved for OUD treatment using data from KASPER (Kentucky All Schedule Prescription Electronic Reporting), the state's prescription drug monitoring program (PDMP).

### 2.4. Preprocessing and Missing Data

We preprocessed our data to normalize geographic and interval groupings and to address missing data. Each geographic region's time series was independently normalized, as different counties or ADDs can have very different populations and event scales, with some having orders of magnitude more SOOs than others. Each county and ADD was individually normalized between 0 and 1, with 1 representing the maximum number of monthly SOOs recorded for that county or ADD. Thus, each geographic grouping had consistent and comparable ranges of values.

The target series contains data for each timestep, but many of the covariates do not. Each external data source covers different time frames; for example, some data sources may only begin as late as 2022, while others may not be updated past 2023 due to lags in data processing or availability. The time series itself cannot be limited only to time periods where all data is present, as this would eliminate a large quantity of data for training and would prevent current predictions from being generated until historical covariate data aligns. We tested three techniques to address missing data: a constant fill value, data imputation, and fitting a simple forecasting model to predict missing data. We created an additional flag as a covariate to each forecasting model that indicates whether the data was initially missing, ensuring that the models can still distinguish between real covariate data and synthetic covariate data. In this use case, missing data can be filled under the assumption that the only timestamps that will contain missing data will be at the beginning or end of the time series.

For simplicity, we first tested using a constant fill value technique, where a single value is used to replace all missing data. We utilized two fill values: one for the beginning of the series (the average of the first year) and one for the end (the average of the last year). We also tested the iterative imputer model[23], which treats each feature as a function of every other feature in the data to determine realistic and appropriate values to replace missing data with. Finally, we tested a simple forecasting model, exponential smoothing[17], to forecast the beginning and end of each covariate series using each covariate individually to predict over the missing timestamps. Although more complex than the simpler methods, it ensured that the missing values were filled with reasonable counts that follow the overall trends of the data. The determination of which method is the most effective is discussed in the Results section.

## 2.5. Training

After preprocessing and aggregating data into the proper format, data was grouped at the monthly level for time, and at the county, ADD, and state levels for geography. Two primary methods for training and evaluation were implemented to evaluate model performance with and without covariates.

The first method uses a 90% training and 10% testing split, to both maximize the amount of training data and adhere to realistic use cases. We wish to enable our community partners to act upon the forecasted trends, which requires forecasting current SOO totals to result in actionable conclusions. To ensure the predictions are relevant for actionable results, we limit the predictions to 3 months beyond the most recent observed data. The second method uses an expanding window for training and evaluation. The model is initially fit on the first two years of data and produces predictions for the next three months. Then, those three months are incorporated into the training set, and the model is refit. It predicts for the following three months, and this cycle continues as the model is continuously refit on new data. By the end, the model has been trained on the entire dataset and evaluated on all but the first two years. This method uses a much more thorough evaluation than the training/testing split, and it makes the best of limited data by incrementally training and evaluating.

In both cases, a single model was trained on a multivariate dataset including data for each grouping. All three grouping levels were combined into a single dataset. Therefore, the dataset used contained 136 target series, consisting of 120 counties, 15 ADDs, and one state level aggregation. Due to the relatively short length of each time series, hyperparameter tuning was not performed for any model to avoid overfitting. Both methods were evaluated primarily using the Root Mean Squared Error (RMSE) to measure accuracy of the forecasts. For the training/testing split, the RMSE was only calculated for the test set. For the expanding window method, the RMSE was calculated for every three-month prediction. These RMSEs were averaged over the last year of data to ensure a fair review of current performance.

## 2.6 Model Comparisons

To compare performance among the nine model types for (1) model predictions at the ADD level, and (2) model predictions at the county level, Friedman's tests were deployed on average RMSEs[24]. For model types with significant Friedman's tests, post-hoc Nemenyi tests were performed to assess significant pairwise differences. P-values < 0.05 were considered significant.

# 3. Results

## 3.1 Geography/Time

The decision of the most appropriate geographical and temporal aggregations to be modeled was decided by analyzing data sparsity levels in the EMS SOO totals for the different subsets of Kentucky's population of approximately 4.5 million people[25]. Table 1 displays the total proportion of timesteps with a value of 0 at each grouping and temporal aggregation level.

**Table 1.** Sparsity of each geographical and temporal aggregation level.

| Grouping | Number of Groupings | Yearly Sparsity | Quarterly Sparsity | Monthly Sparsity | Weekly Sparsity |
|---|---|---|---|---|---|
| State | 1 | 0 | 0 | 0 | 0 |
| ADD | 15 | 0 | 0 | 0.0007 | 0.0439 |
| County | 120 | 0.025 | 0.0825 | 0.2302 | 0.5497 |
| Zip Code | 940 | 0.4718 | 0.6554 | 0.7763 | 0.8922 |
| Tract | 1292 | 0.1128 | 0.3583 | 0.6265 | 0.8723 |
| Block-group | 3474 | 0.3088 | 0.6189 | 0.8193 | 0.9479 |
| Block | 23973 | 0.7509 | 0.9134 | 0.9676 | 0.9919 |

Based on these results, we determined that any geographical region smaller than the county level was likely too sparse for traditional time series forecasting methods to be effective[26], with most temporal intervals having >50% zeroes in those regions. Regions with these levels of sparsity are difficult to accurately predict and evaluate. Larger groupings have more predictable trends, as the effects of randomness are mitigated with larger sample sizes. Sparse groupings are more subject to random and dramatic changes, which are difficult to predict. Forecasting models can also show misleadingly high accuracy for very sparse regions, as the models will only learn to predict 0 for every timestep, which is usually correct.

For the temporal interval choice, the weekly aggregations were too sparse, consisting of >50% zeroes for most geographical regions. Many of the covariates are collected at the monthly level, preventing their usage for weekly forecasting. The monthly intervals were the best option for our forecasting task, as it maximizes the number of timesteps that can be used for training without encountering high levels of sparsity.

### 3.2 Preprocessing and Missing Data

To evaluate which method was most effective at filling missing data and for creating reasonable synthetic data that accurately follows the trends of the real data, we tested multiple covariates grouped at the statewide level. A one-year period of real data was removed from each covariate, and each fill method was used to synthetically replace that period. The RMSE was then calculated between the real data and the synthetic data to determine which method most accurately replaced the real data. Three different example covariates are presented for comparison. We chose exponential smoothing for filling missing data based on both lower RMSEs and visual inspection of the synthetic data.

**Table 2.** Example of comparison of missing data fill methods using RMSE.

| | Constant Fill Value | Iterative Imputation | Exponential Smoothing |
|---|---|---|---|
| Individuals with OUD receiving MOUD (Medicaid) | 0.0534 | 0.0716 | 0.0466 |
| Drug seizures | 0.1393 | 1.5394 | 0.1437 |
| Individuals receiving Buprenorphine for OUD treatment | 0.0754 | 0.1116 | 0.0509 |

### 3.3 Model Training

We tested different combinations of models, training methods, and covariates to evaluate performance. Each model, training method, and covariate were tested independently, and a final RMSE value was produced for each combination by averaging the RMSEs for each region included in the dataset. We present a comparison between models and training types with the RMSEs for each covariate averaged together in Table 3. For fair comparison, the RMSE for the expanding window method is calculated over the same window as the test set. The results are then split across counties, ADDs, and state level aggregations for further comparison of the impact of geographical groupings.

**Table 3.** Comparison of RMSEs between models and training methods, for each grouping level.

| | | County Level | ADD Level | State Level |
|---|---|---|---|---|
| Linear Regression | Train/Test | 0.2248 | 0.1888 | 0.3486 |
| | Expanding Window | 0.2066 | 0.1566 | 0.1926 |
| N-Linear | Train/Test | 0.2031 | 0.1691 | 0.2549 |
| | Expanding Window | **0.1817** | **0.1352** | **0.1233** |
| Temporal Fusion Transformer | Train/Test | 0.2034 | 0.1686 | 0.1853 |
| | Expanding Window | 0.1947 | 0.1517 | 0.1444 |

For all three models, the expanding window method of training and evaluation produces lower RMSEs over the same time period as the standard training/testing split. The simple linear regression model performs the worst, and the N-Linear model slightly outperforms the Temporal Fusion Transformer. Across geographical groupings, the RMSEs are generally lowest for the ADD groupings, and the county level groupings typically have the highest RMSEs.

### 3.4 Covariates

Table 4 shows a comparison of the impact of each covariate when tested independently, where the expanding window method was used and the RMSE was calculated over the last year. A baseline comparison is provided that uses no additional covariates (other than the month and season).

**Table 4.** Comparison of each covariate and its impact on the model.

| | Linear Regression | | N-Linear | | TFT | |
|---|---|---|---|---|---|---|
| Covariates: | County | ADD | County | ADD | County | ADD |
| Baseline | 0.2068 | 0.1575 | 0.1821 | 0.1432 | 0.1996 | 0.1624 |
| Drug Seizures | 0.2056 | 0.1557 | 0.1756 | 0.1345 | 0.1923 | 0.1416 |
| Drug Test Results | 0.2064 | 0.1616 | 0.1925 | 0.1479 | 0.1890 | 0.1403 |
| Intake/Release | 0.2064 | 0.1599 | 0.1820 | 0.1374 | 0.1888 | 0.1427 |
| MOUD/Behavioral Health (Medicaid) | 0.2063 | 0.1551 | 0.1740 | 0.1206 | 0.1863 | 0.1422 |
| Nearby County Trends | 0.2067 | 0.1542 | 0.1752 | 0.1246 | 0.1957 | 0.1628 |
| PDMP Buprenorphine | 0.2090 | 0.1601 | 0.1867 | 0.1420 | **0.1859** | **0.1388** |
| Similar County Trends | 0.2060 | 0.1551 | **0.1722** | **0.1204** | 0.1893 | 0.1617 |
| Social Determinants of Health | 0.2059 | 0.1549 | 0.1862 | 0.1423 | 0.2087 | 0.1475 |
| Treatment Centers | **0.2053** | **0.1529** | 0.1799 | 0.1303 | 0.1956 | 0.1607 |
| Weather | 0.2080 | 0.1561 | 0.1927 | 0.1521 | 0.2153 | 0.1784 |

Additionally, the state level forecasting performance varied widely depending on the covariate, from a minimum of 0.0725 RMSE (Medicaid) up to 0.2628 (baseline) for the N-Linear model. The county level generally had the lowest variation in RMSE across covariates, and the linear regression model also had very small differences.

### 3.5 Model Comparison

To compare overall model performance across different covariate sets, we analyzed the differences between three primary settings: using no covariates, using some commonly available covariates, and using all covariates. Each of the three model architectures was trained and evaluated using the expanding window method for each of these three settings. The 'no covariates' setting is a baseline version where no covariates other than the month and season are included. The 'all covariates' setting includes every covariate described above at once. The 'common covariates' include a subset of covariates we decided upon based on both performance and availability (similar county trends, nearby county trends, and SDOH). These are the most accessible and easy to use covariates for other researchers, as they do not require any additional non-publicly available data sources.

We performed the Friedman test to compare the output from each of these model configurations. We compared nine total model types: three different architectures (linear regression, N-Linear, TFT) and three different covariate settings (none, common, all). This test was performed twice: once for the RMSEs at the county level and once for the RMSEs at the ADD level, since the test assumes independence of observations. At the county level, the Friedman test produced a test statistic of 240.9156 and p-value < 0.0001. At the ADD level, the test statistic was 52.3022 and the p-value < 0.0001. Since both tests demonstrated significant differences in model architecture and covariate set performance, Nemenyi post-hoc tests were then performed to make pairwise comparisons. As an example, subsets of the post-hoc test results are provided below, highlighting significant differences between base models and each covariate setting.

**Table 5.** P-values from Nemenyi post-hoc test for county and ADD RMSEs. Shading indicates comparisons of same model architecture.

| Grouping Level | | TFT None | Regression None | N-Linear None | N-Linear Common | N-Linear All |
|---|---|---|---|---|---|---|
| County | TFT None | 1 | 0.9999 | 0.0001 | <0.0001 | 0.9794 |
| | Regression None | 0.9999 | 1 | <0.0001 | <0.0001 | 0.9863 |
| | N-Linear None | 0.0001 | <0.0001 | 1 | 0.0020 | <0.0001 |
| | N-Linear Common | <0.0001 | <0.0001 | 0.0020 | 1 | <0.0001 |
| | N-Linear All | 0.9794 | 0.9863 | <0.0001 | <0.0001 | 1 |
| ADD | TFT None | 1 | 0.8983 | 0.0371 | 0.0001 | 0.9946 |
| | Regression None | 0.8983 | 1 | 0.6823 | 0.0301 | 0.9998 |
| | N-Linear None | 0.0371 | 0.6823 | 1 | 0.8711 | 0.3219 |
| | N-Linear Common | 0.0001 | 0.0301 | 0.8711 | 1 | 0.0046 |
| | N-Linear All | 0.9946 | 0.9998 | 0.3219 | 0.0046 | 1 |

At the county level, significant differences are observed between all three covariate settings for the N-Linear model. Additionally, there are significant differences between the N-Linear, TFT, and Regression models. Fewer ADD comparisons were significant, which may be explained by the smaller sample size of 15 regions instead of 120 counties. The N-Linear model with common covariates was ranked the highest in performance, with an average RMSE across all regions of 0.1598, compared to 0.1740 for no covariates and 0.2013 for all covariates. Additionally, the N-Linear model ranked overall higher than the linear regression or TFT models, which achieved RMSEs of 0.1952 and 0.2038 respectively with no covariates.

Figure 1 shows two plots displaying the performance of the expanding window N-Linear model. In plot (A), the statewide level predictions are shown for the model using the common covariates subset. The predictions begin after two years of training and improve over time as more data is used for training. Plot (B) further proves this, showing the decrease in RMSE across all regions when evaluated in the expanding window process.

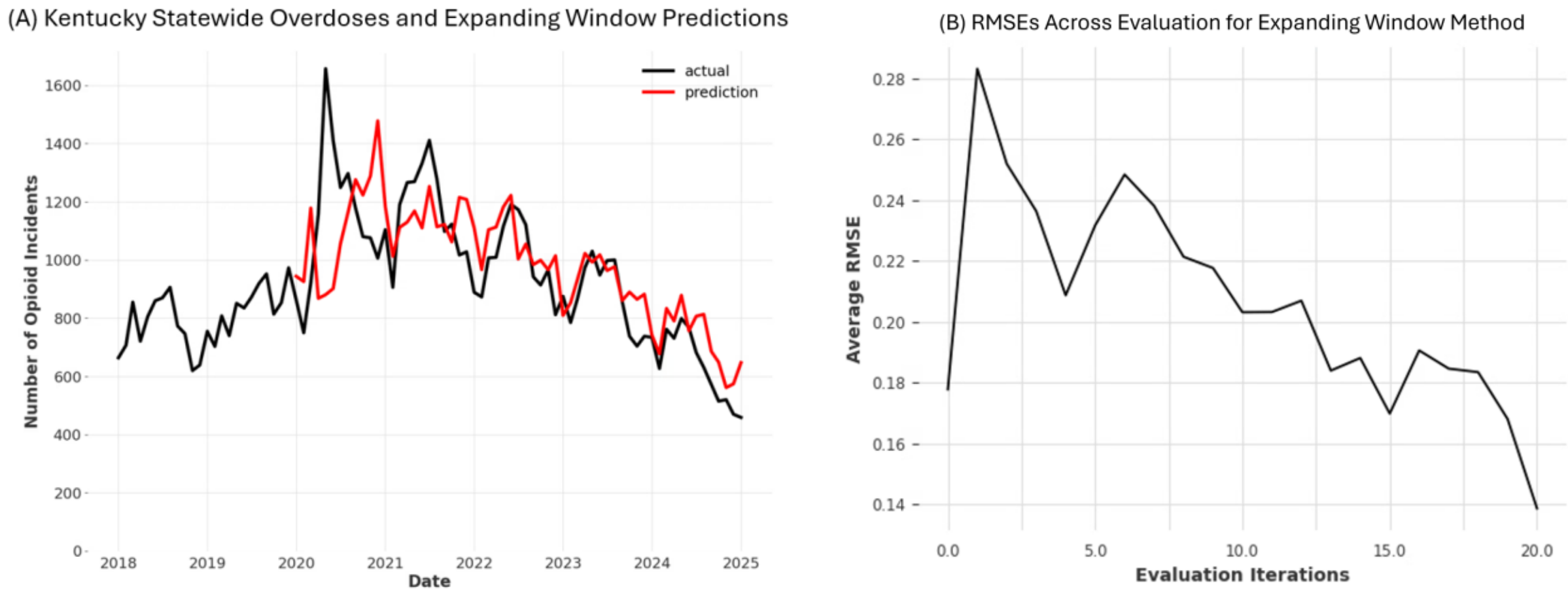

**Figure 1.** Statewide predictions (A) and RMSEs (B) from the expanding window method.

## 4. Discussion

Tables 3 and 4 show that the choice of geographical aggregation has a large influence on prediction accuracy. Every model has consistently lower errors at the ADD level than at the county level, which is due to the sparsity of incidents in many smaller and less populated counties that may only have a few SOOs every year. By combining these areas into ADDs, the number of SOOs increases at each timestep, and this allows the models to better understand the trends of the data, as the larger scale will be more predictable and less random. In practice, this result shows that the ADD aggregation level should be used for smaller counties, while the county level aggregation may be used for larger counties to provide more resolution for urban populations. State-level predictions seem more variable but can achieve lower errors than both county and ADD regions. Figure 1 shows the overall statewide trend, and a clear yearly repeating pattern can be observed. This pattern is most apparent in large regions, likely contributing to the increased accuracy observed. State-level predictions can be useful for demonstration of overall forecasting abilities but serve less practical purpose than predictions for smaller regions; allocation of state-funded resources such as naloxone would require data per county[3,4].

Many of the covariates also have bigger impacts at the ADD level compared to the county level because covariates also have sparsity problems in smaller regions. In bigger regions, the covariates can contribute more information to the model, further improving accuracy. Some of the external covariates discussed here have strong positive impacts on the model, such as Medicaid claims data and DOC intake/release measures. Others, such as urine drug testing results, provide little improvement to performance despite strong correlations with SOOs. This may be because, while these covariates' trends align with the target series, they hold little predictive power for future forecasts. Several of the strongest covariates do not require these additional data sources. Both nearby county trends and similar county trends can be extracted from the target variable itself, and the social determinants of health come from a publicly available dataset. We have shown that a model incorporating only these covariates is still significantly better than a baseline model. Thus, the solutions discussed here can be adapted to other states or regions without requiring extensive data collection. The inclusion of all covariates proved to be ineffective, but in the future, we will work to comprehensively test different combinations of covariates to determine the most effective configurations for forecasting.

We also note that RMSE is not a perfect error metric, especially when determining how useful the predictions are in practice. Some of the smallest counties have the lowest errors because SOOs are so rare; this causes the model to predict 0 for the entire forecast window, which is highly likely to be right. However, a model that predicts no SOOs will ever occur in a county is not useful because rare events do eventually occur. Other metrics, such as Mean Absolute Error, are more robust to skewed distributions and can provide a clearer picture of model performance in these cases. Mean Absolute Scaled Error can also be used to judge how effectively a model improves upon a baseline prediction value, which can be helpful for regions where most of the values are the same over time. Grouping into ADDs also helps to resolve this problem by ensuring each time series is active enough to produce valuable, diverse forecasts. We chose to primarily use RMSE for comparison because it more harshly punishes large errors than other metrics, and for our use case, large prediction errors can have important consequences for stakeholders that we wanted to avoid.

The best performance in testing was from the N-Linear model, which achieved the lowest error in nearly all cases. This model was simple enough to prevent overfitting on the target series, while in-depth enough to learn trends across multiple series. The linear regression model performed the worst and seemed unable to effectively utilize the covariates. This can be seen in Table 4, where the RMSE varies only slightly between different covariates. The TFT performed slightly worse than the N-Linear model, potentially because the TFT is too large and complex for this use case. With monthly data beginning in 2018, each time series only has around 80 timesteps, which is a relatively small amount of data for time series analysis. The inclusion of multiple different series helps to offset this, but large transformer models are generally best suited for large datasets which limits their utility for our use case.

We have partnered with our state government and have identified key stakeholders responsible for public health policy and allocation of resources related to overdose. Our predictions are used in publicly available dashboards that allow our partners and stakeholders to view historical data and forecasts for their own regions for decision making. These dashboards are updated every month with new three-month predictions, ensuring users can stay up to date on the latest trends in SOOs, which may assist in the allocation of resources such as Naloxone, which is a highly effective tool preventing fatal opioid overdoses (re-cite the naloxone paper #4 in ref. list). These predictions are generated using the N-Linear model and a combination of different covariates, including the MOUD/Behavioral Health measures from Medicaid, DOC intake/release measures, the PDMP buprenorphine measure, nearby county trends, and similar county trends. These covariates were chosen based on a combination of both performance and availability. The publicly available dashboards can be found here: https://rador-ky.uky.edu/public_dashboards.

## 5. Conclusion

We found that suspected opioid overdoses can be forecasted within a reasonable degree of error across the state of Kentucky. It is best to aggregate this data at larger district levels; smaller groupings, such as county level, are frequently very sparse except for the large urban areas. A variety of covariates extracted from external data sources were utilized in this work, the inclusion of which had positive impacts on forecasting performance. Multiple models were tested and the N-Linear model, which uses a simple one-layer architecture, had the best performance with these covariates. These results could be used by state agencies to determine when and where opioid overdoses can be expected to increase or decrease. This information will allow resources and personnel to be distributed effectively, with the goal of reducing overdoses while providing more efficient treatment when they do occur. In the future, more data sources, methods, models, and covariates will continue to be tested to improve performance.

## 6. Acknowledgement

This research was supported by the National Institute on Drug Abuse (NIDA) of the National Institutes of Health under award number R01DA057605. The content is solely the responsibility of the authors and does not necessarily represent the official views of the National Institutes of Health.

The authors thank the Kentucky Office of Vital Statistics, Kentucky Department for Public Health, Kentucky Medical Examiners' Office, the Kentucky Board of Emergency Medical Services, the Kentucky State Police, and Millenium Health for their support for this study and providing data.